\documentclass[sigconf]{acmart}
\usepackage{multirow}

\usepackage{algorithm}
\usepackage{algorithmic}
\usepackage{pifont}
\usepackage{tablefootnote}

\AtBeginDocument{%
 }
  
\copyrightyear{2024}
\acmYear{2024}
\setcopyright{rightsretained}
\acmConference[SIGIR '24]{Proceedings of the 47th International ACM SIGIR Conference on Research and Development in Information Retrieval}{July 14--18, 2024}{Washington, DC, USA}
\acmBooktitle{Proceedings of the 47th International ACM SIGIR Conference on Research and Development in Information Retrieval (SIGIR '24), July 14--18, 2024, Washington, DC, USA}
\acmDOI{10.1145/3626772.3657760}
\acmISBN{979-8-4007-0431-4/24/07}

\makeatletter
\gdef\@copyrightpermission{
  \begin{minipage}{0.3\columnwidth}
   \href{https://creativecommons.org/licenses/by/4.0/}{\includegraphics[width=0.90\textwidth]{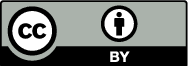}}
  \end{minipage}\hfill
  \begin{minipage}{0.7\columnwidth}
   \href{https://creativecommons.org/licenses/by/4.0/}{This work is licensed under a Creative Commons Attribution International 4.0 License.}
  \end{minipage}
  \vspace{5pt}
}
\makeatother

\begin{document}

\title{IM-RAG: Multi-Round Retrieval-Augmented Generation Through Learning Inner Monologues}


\author{Diji Yang}
\email{dyang39@ucsc.edu}
\affiliation{%
  \institution{University of California Santa Cruz}
  \city{Santa Cruz}
  \country{USA}
}

\author{Jinmeng Rao}
\email{jinmengrao@gmail.com}
\affiliation{%
  \institution{Mineral.ai}
  \city{Mountain View}
  \country{USA}
}

\author{Kezhen Chen}
\authornotemark[1]
\email{kzchen0204@gmail.com}
\affiliation{%
  \institution{Together AI}
  \city{Mountain View}
  \country{USA}
}

\author{Xiaoyuan Guo}
\authornotemark[1]
\email{xiaoyuanguo@google.com}
\affiliation{%
  \institution{Google}
  \city{Mountain View}
  \country{USA}
}

\author{Yawen Zhang}
\email{yawenz1129@gmail.com}
\affiliation{%
  \institution{Mineral.ai}
  \city{Mountain View}
  \country{USA}
}

\author{Jie Yang}
\authornote{Work done at Mineral.ai.}
\email{jie@cybever.ai}
\affiliation{%
  \institution{Cybever}
  \city{Mountain View}
  \country{USA}
}

\author{Yi Zhang}
\email{yiz@ucsc.edu}
\affiliation{%
  \institution{University of California Santa Cruz}
  \city{Santa Cruz}
  \country{USA}
}
\renewcommand{\shortauthors}{Diji Yang et al.}


\begin{abstract}
Although the Retrieval-Augmented Generation (RAG) paradigms can use external knowledge to enhance and ground the outputs of Large Language Models (LLMs) to mitigate generative hallucinations and static knowledge base problems, they still suffer from limited flexibility in adopting Information Retrieval (IR) systems with varying capabilities, constrained interpretability during the multi-round retrieval process, and a lack of end-to-end optimization. To address these challenges, we propose a novel LLM-centric approach, \textbf{IM-RAG}, that integrates IR systems with LLMs to support multi-round RAG through learning Inner Monologues (IM, i.e., the human inner voice that narrates one's thoughts). During the IM process, the LLM serves as the core reasoning model (i.e., \textit{Reasoner}) to either propose queries to collect more information via the \textit{Retriever} or to provide a final answer based on the conversational context. We also introduce a \textit{Refiner} that improves the outputs from the \textit{Retriever}, effectively bridging the gap between the \textit{Reasoner} and IR modules with varying capabilities and fostering multi-round communications. The entire IM process is optimized via Reinforcement Learning (RL) where a \textit{Progress Tracker} is incorporated to provide mid-step rewards, and the answer prediction is further separately optimized via Supervised Fine-Tuning (SFT). We conduct extensive experiments with the HotPotQA dataset, a popular benchmark for retrieval-based, multi-step question-answering. The results show that our approach achieves state-of-the-art (SOTA) performance while providing high flexibility in integrating IR modules as well as strong interpretability exhibited in the learned inner monologues.
\end{abstract}

\ccsdesc[500]{Information systems~Question answering}
\ccsdesc[500]{Information systems~Language models}

\keywords{retrieval augmented generation, inner monologue, large language models, question answering, multi-round retrieval}

\maketitle

\section{Introduction}
\begin{figure*}[tbh]
    \centering
    \includegraphics[width=0.90\linewidth]{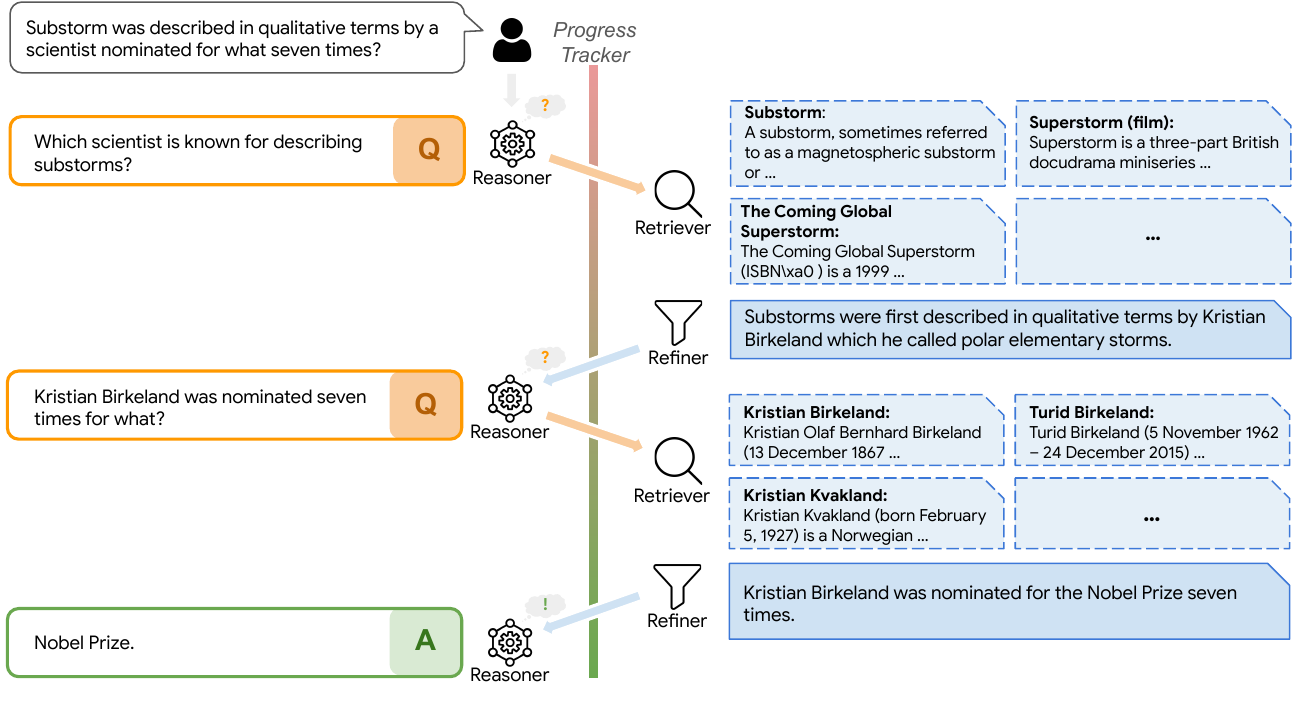}
    \caption{
    The Inner Monologue (IM) process in IM-RAG. For users posed questions, the \textit{Reasoner} first determines if it has enough information to provide an answer. If not, it acts as a \textit{Questioner}, proposing a query to request more information. The query is then directed to the \textit{Retriever}, which searches for relevant documents in the knowledge source. Subsequently, the \textit{Refiner} refines the retrieved documents to highlight the most pertinent information, which is then returned to the \textit{Reasoner}. This iterative process may continue over multiple rounds until the \textit{Reasoner} believes it has gathered enough information, at which point it becomes an \textit{Answerer} and generates a final answer. This IM process provides valuable insights into the reasoning process, enabling humans to understand how the system arrived at its conclusions.
    }\label{fig:teaser}
\end{figure*}

Large Language Models (LLMs) have demonstrated impressive capabilities in language understanding and generation~\citep{palm, ouyang2022training, touvron2023llama}; however, there are two major challenges: generative hallucination ~\citep{welleck2019neural} and static knowledge~\citep{komeili2021internet}. While LLMs possess a deep understanding of human language and can generate creative responses, they lack the ability to verify facts or access up-to-date information ~\citep{mialon2023augmented, ai2023information}. To mitigate such issues, integrating Information Retrieval (IR) systems with LLMs has become an increasingly promising direction. IR systems complement LLM by retrieving timely and relevant information, enhancing the factuality of responses. The synergy between LLMs and the IR systems -- Retrieval Augmented Generation (RAG)~\citep{shuster2021retrieval, mialon2023augmented} improves the ability of LLMs and powers generative AI products like ChatGPT, Bard, and Bing, showcasing the power and future potential of the combining IR systems and LLMs for more accurate and reliable responses.

There are two typical paradigms to improve RAG systems: the joint training approach v.s. training different components separately. The first paradigm involves joint training of LLMs and retrievers on knowledge-intensive tasks, enhancing retrieval capabilities of language models~\citep{izacard2022few}. For example,~\citet{guu2020retrieval} did joint training of LLM and a retriever's semantic embedding, and their approach has shown promising results. However, it lacks interpretability because the communication between LLMs and retrievers relies on complex deep-learning gradient propagation and cross-attention between IR embedding models and LLMs. Furthermore, this training approach is very computationally expensive, and it's very hard or expensive to retrain the retriever's semantic embedding as LLMs change or learn.
The second paradigm improves LLM and/or IR engines separately. Most prior work in this paradigm focuses on improving LLM (LLM-centric), either through prompting or fine-tuning LLM parameters ~\citep{ram2023context, lazaridou2022internet, nakano2021webgpt}. The prompting-based approach provides simplicity and flexibility without incurring extra training costs and allows the integration of black-box LLMs and search engines through API calls. However, it suffers from the lack of end-to-end optimization of the whole system. For example, efforts spent on improving LLM search query rewriting/generation module may not lead to better retrieval performance, as the improvement is not well tailored for the specific search engine used. Besides, a static LLM generation module may not perform well when fed with both relevant and irrelevant documents. In contrast, a training-based approach collects and utilizes human-annotated interaction records between LLMs and IR modules, and then uses them to supervise LLMs in learning how to better utilize and interact with IR modules. Although this approach has shown better performance than the prompting-based approach on simple image-to-text retrieval-based visual question-answering tasks ~\citep{liu2023llava}, it requires a significant amount of labeled training data as well as substantial training costs. For complex problems that require multi-step reasoning and multi-round retrieval, training data with human-labeled multi-round search records can be expensive to collect, and the effectiveness of their method is unclear. In this work, we mainly focus on improving the LLM-centric paradigm, considering its performance, flexibility, and interpretability.

Recently, IMMO~\citep{yang2023tackling} trained an LLM and a vision-language mode to have Inner Monologues (i.e. Question-Answering (QA) dialogues), and their results show the learned IM does explicit multi-step reasoning, performs well on complex visual Question Answering problems, meanwhile explainable. 

Motivated by IMMO, we adapt the concept of IM to RAG to enable LLMs to do multi-round retrieval, as we believe learning IM could also be beneficial for the communication and collaboration between LLM and IR modules. Prior cognitive science studies suggest that human Inner Monologue encompasses a broader spectrum of mental processes beyond QA dialogues, including abstract concepts and symbolic representations~\citep{vygotsky1987thinking,innerspeech}. Thus, in this paper, we extend IM communication beyond the format of QA dialogues in natural language, and further generalized IM to involve more formats that are more appropriate for RAG systems (e.g., ranking results and returning scalar scores). This leads to a novel LLM-centric framework \textbf{IM-RAG} that integrates LLMs with IR to support context-aware multi-round interactive retrieval through learning IM.
In our framework, LLM (i.e., \textit{Reasoner}) acts as the mastermind of IM-RAG, switching between two crucial roles during the multi-round communication smoothly. When additional information is needed, it becomes a \textit{Questioner}, crafting new queries based on the conversational contexts to acquire more relevant documents from \textit{Retriever} (i.e., a search engine); when enough information is gathered, it automatically transitions to an \textit{Answerer}, summarizes search results for the original user query, and sends the final responses to the user. 
To better adapt a search engine to an LLM, we add a \textit{Refiner} component after the \textit{Retriever}. 
This component learns to refine retrieved documents (e.g., reranking or reformatting) to meet the needs of LLM. This helps the LLM's reasoning process and facilitates the interaction with \textit{Retriever} as it bridges the gap between LLMs and retrievers. With a \textit{Refiner} as a learnable adapter, one can switch or add more IR modules without worrying much about the change of IR module capabilities and output formats. \textit{Progress Tracker} for LLM is introduced to track the multi-round retrieval progress, so that LLM can switch its roles from questioner to answerer. We use RL to optimize the IM interaction between LLM and \textit{Retriever} with multi-round retrieval progress as reward signals. Figure~\ref{fig:teaser} shows one example of how our
IM-RAG system solves complex question-answering problems through multi-round retrieval. We summarize our contributions as follows:

\begin{itemize}
    \item Inspired by IMMO, we introduce a novel approach, \textbf{IM-RAG}, that connects LLMs and IR modules for context-aware multi-round RAG through learning IM. The IM learning process can be optimized via RL without intermediate human annotations. The learning process enables the key components of a RAG system (query generation, results ranking, answer generation, etc.) to be trained to match the capability of other components. Thus, the whole RAG system is optimized. 
    \item Our work offers a solution that provides flexibility in adopting IR modules and LLMs with varying capabilities, as well as interpretability for multi-round retrieval.
    \item We demonstrate the efficacy of our approach on the HotPotQA dataset~\citep{yang2018HotPotQA}, a popular knowledge-intensive multi-hop question-answering dataset, and our approach achieves SOTA performance.
\end{itemize}

\begin{figure*}[tbh]
    \centering
    \includegraphics[width=0.75\linewidth]{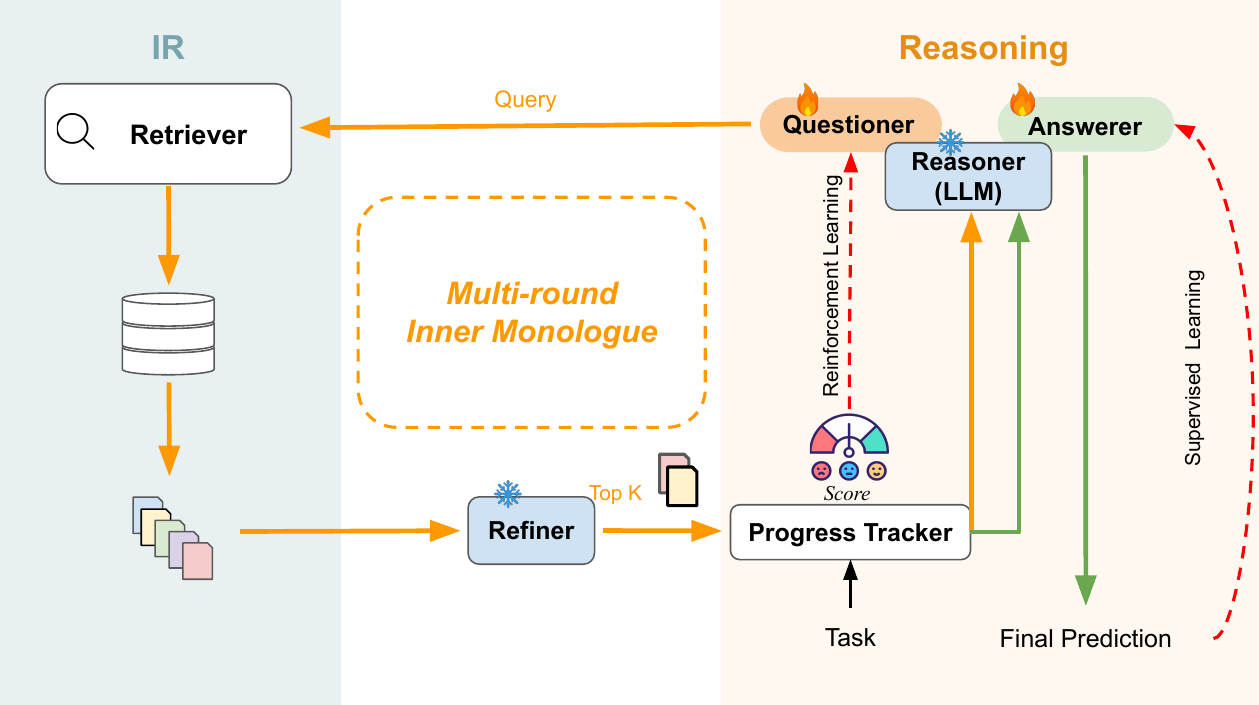}
    \caption{Overview of IM-RAG framework. It involves four main components: a \textit{Reasoner}, a \textit{Retriever}, a \textit{Refiner}, and a \textit{Progress Tracker}. The \textit{Reasoner} is responsible for core reasoning, switching its role between \textit{Questioner} (learning to propose queries to request relevant documents via the \textit{Retriever}) and \textit{Answerer} (learning to predict a final answer based on the conversational context). The \textit{Refiner} improves the retrieved documents via rephrasing or reranking and passes the top-k highlighted documents to both the \textit{Progress Tracker} for predicting progress scores and the \textit{Reasoner} for further reasoning. The training of \textit{Questioner} happens during the RL stage, where the progress scores are used as rewards. The training of \textit{Answerer} happens during the SFT stage, where the original questions, learned IM with refined top-k documents at each turn, and ground truth answers are used as finetuning examples.
    }\label{fig:overview}
\end{figure*}

\section{Related works}
\label{sec:related}

\paragraph{Retrieval-Augmented Generation for LLMs}
Language models often face challenges such as generating hallucinations or being constrained by static knowledge bases. RAG has been identified as a potential solution to tackle these challenges, offering reliable grounding and the flexibility to access various external knowledge bases. One paradigm of RAG is to jointly train language models and retrievers on knowledge-intensive tasks~\citep{guu2020retrieval,lewis2020retrieval,izacard2022few,lin2023ra}. For example, REALM~\citep{guu2020retrieval} models retrieved documents as latent variables and jointly pretrains a BERT-style language model and a neural retriever through Masked Language Modeling (MLM). Atlas~\citep{izacard2022few} demonstrates that joint training can also bring strong few-shot learning capabilities on a wide range of knowledge-intensive tasks. RA-DIT~\citep{lin2023ra} proposes a dual instruction tuning method to retrofit language models with retrieval capabilities and achieves SOTA performance on many knowledge-intensive zero-shot and few-shot learning benchmarks.
With the rise of LLMs, building LLM-centric systems emerges as another popular paradigm of RAG, where an LLM acts as a core reasoning model, and other models and tools (including retrievers such as search engines and neural retrievers) are integrated with the LLM through prompting or training. For example, HuggingGPT~\citep{shen2023hugginggpt} and Chameleon~\citep{lu2023chameleon} prompt LLMs with tool descriptions and use examples to accomplish various complex reasoning tasks by composing various tools. Though these prompt-based methods offer flexible plug-and-play solutions, they are hard to optimize end to end. Other works, such as ToolFormer~\citep{schick2023toolformer}, train LLMs on filtered and sampled API calls to teach LLMs how to use a variety of tools. These training-based methods can be supervised while requiring a large number of training data and providing limited interpretability for multi-round retrieval. Our work focuses on enhancing the multi-round retrieval capabilities of LLM-centric systems through IM learning, which can be optimized end-to-end without heavy training data curation costs while providing high flexibility and interpretability.

\paragraph{Question Answering}
The evolution of Question-Answering (QA) research, particularly within the realm of information retrieval, has been significantly influenced by initiatives like the Text Retrieval Conference (TREC) QA track in early 2000. Traditional approaches of open domain QA usually include a retriever that finds relevant documents and a reader that processes retrieved documents to generate answer candidates. Extensive research has been done to study how to improve retriever-based, such as iterative approaches that sequentially update search queries at each iteration. Most of those approaches do not change the retriever or the reader. Recently,~\citet{zhu2021adaptive} models the iterative retrieval and answer process as a partially observed Markov decision process, carefully designed actions and states of the agents, and trained each component of the system.~\citet{ma-etal-2023-chain} proposes to chain together carefully design skills or modules, each specialized in a specific type of information processing task, for question answering, and one skill is retrieval based on a query expanded with the previous-hop evidence for multi-round retrieval. Our proposed research is motivated by the success of prior research on iterative retrieval, while we are more focused on enhancing the ability of large-scale language models, and we proposed a novel iterative retrieval solution that's more general and explainable based on the strength of LLMs. 

\paragraph{Inner Monologue}
Recent studies have demonstrated the significant potential of LLM-centric systems in reasoning, planning, fact-checking, and knowledge management through carefully crafted chain-of-thought prompts, facilitating multi-agent collaboration~\cite{huang2022inner, wei2022emergent, yang2023gpt4tools}. As a cognitive process, Inner Monologue (i.e., self-talk conducted through the internal stream of thoughts) has recently been recognized as an efficient prompting strategy for LLM-centric systems~\cite{cherney2023everything,huang2022inner,yang2023tackling,wang2023adapting}. 
For example, by leveraging environmental feedback, Huang et al.~\cite{huang2022inner} apply IM into an LLM-centric system to enable grounded closed-loop feedback for robot planning and reasoning.
Zhou et al.~\citep{zhou2023think} design and add IM to enable LLMs to better understand and use communication skills.
IMMO~\citep{yang2023tackling} proposes that natural language QA dialogues between an LLM and a Vision-Language Model (VLM) can serve as a form of IM, which can be further optimized end-to-end via RL. However, this QA-based IM is restrictive, as it only facilitates interactions among models capable of processing and responding in QA formats. In the field of IR, many traditional IR modules' inputs and outputs may not form QA pairs or even natural language.
In this work, we further extend the IM within LLM-centric systems to any form of communication between the "\textit{Reasoner}" and "\textit{Retriever}" (e.g., lists of text chunks, ranking results, or scalar scores), either structured or unstructured, to provide high flexibility for communication and room for optimization. A "\textit{Refiner}" is added after the "\textit{Retriever}" to refine any form of output into a desired format and length for LLMs. Our approach is anticipated to be a versatile framework that facilitates collaboration between components in LLM-centric systems.

\section{Methodology}
In this section, we first briefly review the IMMO process~\citep{yang2023tackling}, which shares a similar learning framework with our approach. Then, we present IM-RAG as well as the rationales behind the design.

\subsection{Review of IMMO}

IMMO tackles the commonsense visual question-answering tasks by leveraging the LLM's rich common-sense knowledge in conjunction with VLM's image-understanding capabilities. During the learning stage, the LLM engages in a dialogue with VLM in natural language format, which is the IM process in the system. After multiple turns of conversation, the LLM gathers enough information and provides a final answer. The whole IM process is optimized through Proximal Policy Optimization (PPO)~\citep{schulman2017proximal} that is based on the correctness of the final answer and penalized by the Kullback–Leibler (KL) divergence between the updated and the initial policy~\citep{jaques2017sequence}. This approach does not require human-annotated multi-round conversations for RL and only uses the correctness of the final answer as reward signals. Despite that IMMO achieves impressive performance, the lack of mid-step rewards makes it difficult to optimize the behavior at each step during the overall multi-step reasoning process. 
Additionally, the QA-based IM used in IMMO can be restrictive. It is important to recognize that in an LLM-centric system, various interactions, such as communications with retrievers, don't always rely on natural language dialogues. In our work, we broaden the form and use of IM to include information retrieval. Our approach introduces mid-step rewards to provide more detailed and precise feedback at each step during the RL process, improving the system's capability in the multi-round interactive retrieval.

\subsection{The IM-RAG Approach}
\label{sec:approach} 
IM-RAG, as depicted in Figure~\ref{fig:overview}, is an LLM-centric system, which consists of four components: a \textit{Reasoner}, a \textit{Retriever}, a \textit{Refiner}, and a \textit{Progress Tracker}. The components are connected through multi-round Inner Monologues. Below we first illustrate the design of each component, then describe the training process of our approach.

\subsubsection{Reasoner}
\label{sec:reasoner}
As shown in Figure~\ref{fig:overview}, the \textit{Reasoner} serves as the core reasoning component in the IM-RAG framework with two key responsibilities: (1) Questioning: crafting search queries to acquire relevant documents iteratively through IR; (2) Answering: providing the final answer to the initial question based on the multi-round interaction between the \textit{Reasoner} and the \textit{Retriever} (i.e., Inner Monologues within IM-RAG). For these two responsibilities, we introduce two distinct parameter-efficient adapters to specialize each capability during the learning process. Specifically, we added two LoRA~\citep{hu2021lora} adapters to the same base LLM, namely \textit{Questioner} and \textit{Answerer}. We first train the \textit{Questioner} through its multi-round IM with the \textit{Retriever} via reinforcement learning. During this RL stage, the \textit{Questioner} learns how to decompose a complex task (e.g., a question that requires multi-step retrieval and reasoning) into a series of simpler sub-queries. The sub-queries depend on the previous communication context, which can include the sub-query and the retrieved documents in the previous step, as well as the original question. We then train the \textit{Answerer} through Supervised Fine-Tuning (SFT) to directly answer the original question. During the SFT stage, the \textit{Answerer} leverages the IM learned from the RL stage and provides a correct answer. The detailed training strategies of two adapters are illustrated in section~\ref{sec:questioner}~and~\ref{sec:answerer}, respectively.

\subsubsection{Retriever}
\label{sec:retriever}
As shown in Figure~\ref{fig:overview}, the purpose of the \textit{Retriever} component in the IM-RAG is to accurately retrieve relevant documents given search queries from the \textit{Reasoner} during the IM process. The specific architecture of the \textit{Retriever} and its knowledge resources can be flexible depending on various tasks or datasets. Conceptually, most existing search engines, dense retrievers, or matching algorithms can be directly adopted into the IM-RAG framework as the \textit{Reasoner}. There are two reasons behind this design: (1) all the components in IM-RAG are fully decoupled, which makes IM-RAG an efficient plug-and-play solution; (2) the \textit{Refiner} component (introduced below) is able to refine a variety of outputs from different IR modules into the content of a desired format and length, which gives more freedom in the selection of the \textit{Retriever}.

\subsubsection{Refiner}
\label{sec:refiner}
As illustrated in Figure~\ref{fig:overview}, we introduced a \textit{Refiner} component in the IM-RAG to enhance the inner monologue process, particularly the multi-round conversations between the \textit{Reasoner} and the \textit{Retriever}. The \textit{Refiner} serves as a post-processor for the \textit{Retriever}'s outputs. Its introduction is driven by two primary motivations: First, the outputs from various IR modules differ in format and length, which might not be ideally suited as contextual prompts for LLMs. The \textit{Refiner} addresses this by rephrasing and standardizing these outputs into concise, well-formatted passages. Second, the varying capabilities of different IR modules can lead to unfiltered or unranked results, which can limit their utility. The \textit{Refiner} improves these results by reranking and filtering, making sure only the important information stands out. In essence, the \textit{Refiner} provides flexibility to the choice of IR modules and ensures their compatibility with the \textit{Reasoner}, effectively bridging the gap between the \textit{Retriever} and the \textit{Reasoner} and streamlining the IM process.

\subsubsection{Progress Tracker}
\label{sec:progress_tracker}
RL algorithms such as PPO are inherently plagued by optimization inefficiencies when the search space is huge~\citep{schulman2017proximal}. One way to mitigate these inefficiencies is by providing well-designed mid-step rewards during the multi-round process~\citep{lightman2023let, uesato2022solving}. Thus, we introduce a \textit{Progress Tracker} component in IM-RAG to provide a reward score based on retrieval progress at each turn. When the accumulated score exceeds a certain threshold, it indicates that the \textit{Reasoner} has acquired sufficient information and should give a final answer. In practice, the scoring design of the \textit{Progress Tracker} can be flexible, varying across different tasks, retrievers, and datasets. This flexibility may include a neural reward model~\citep{ouyang2022training} or a discrete reward function~\citep{yang2023tackling}. In IM-RAG, we introduce a soft distance score design based on cosine similarity, which provides robust reward signals while maintaining simplicity.

Denote the top 1 passage from \textit{Refiner} at $i$-th turn is $pr_i$, and $\{p_1, p_2, ..., p_n\}$ be list of golden support passages ($SP$), where $n$ is the length of $SP$. The closest passages to $pr_i$ can be found by cosine similarity. For brevity, the $cos$ function shown in Equation~\ref{eq:find_closest}~and~\ref{eq:find_distance} includes the operation of encoding passage into embedding space. 

\begin{equation}
p_{closest} = \underset{p \in {SP}}{\mathrm{argmax}} \,\, \cos(pr_i, p) 
\label{eq:find_closest}
\end{equation}

\begin{equation}
d_i = 1- {cos}(pr_i, p_{closest}) 
\label{eq:find_distance}
\end{equation}

The distance score $d_i$ indicates the quality of $pr_i$, which is bounded with the query $q_i$. Since $p_{closest}$ is considered to have been (attempted to be) retrieved, it will be removed from $SP$. By updating the list of passages that haven't been retrieved yet, dependencies are set between IM turns. The distance score of subsequent turns will partially depend on all preceding actions.

\begin{algorithm}[tbh]
\begin{flushleft} 
\textbf{Dataset}: (Question $Q$, Support passages $SP$, Ground Truth $G$) tuples \\
\textbf{Inner Monologue:} an empty list $IM$ to store inner monologues \\
\textbf{Questioner}: LoRA weights of a pre-trained large language model \\
\textbf{Retriever}: a pre-defined searching system \\
\textbf{Z}: pre-defined training epoch \\
\end{flushleft}
\caption{Reinforcement Learning for Questioner training}
\label{alg:RL}
\begin{algorithmic}[1]
\FOR{epoch = 1 to Z}
    \STATE Define the $Questioner$ as the active model $\mathcal{M}$
    \STATE Sample ($Q$, $SP$, $G$) from the dataset

    \WHILE{Questioner $ \gets $ \{Eq.~\ref{eq:switch_qa_func}\}}    \label{alg:line-IM}
        \STATE $q \gets \mathcal{M}(Q, IM)$
        \STATE $p_s \gets Retriever(q, D)$
        \STATE $p_r \gets Refiner(q, p_s)$  
        \STATE $IM = IM+q+p_r$
        \STATE $p_{closest} \gets$  \label{alg:line-PT}
        \COMMENT{Eq.~\ref{eq:find_closest}}
        \STATE $d \gets $
        \COMMENT{Eq.~\ref{eq:find_distance}}
        \STATE Remove $p_{closest}$ from $SP$     \label{alg:line-PT-end}
    \ENDWHILE     \label{alg:line-IM-end}
    
    \STATE $A_{f} = \mathcal{M}(Q, IM)$
    \STATE $\mathcal{R}$ $\gets$ \COMMENT{Eq.~\ref{eq:final-reward}}
    \STATE PPO updates $\mathcal{M}$ using Reward $\mathcal{R}$
\ENDFOR
\end{algorithmic}
\end{algorithm}

\subsubsection{Questioner Training}
\label{sec:questioner}
The overall training procedure is shown in Algorithm~\ref{alg:RL}. For a given question $Q$, we use the \textit{Questioner} to generate the queries. The training starts with initializing the \textit{Questioner} LoRA as the activate model $\mathcal{M_0}$, an empty list to store the inner monologues $IM$, and the data sample of (question, golden support passages list, ground truth answer) tuple as ($Q$, $ SP$, $G$) from the dataset. 
The multi-round IM process starts from \textit{Progress Tracker} receives the question, as described in the Line~\ref{alg:line-IM}.
The Questioner first generates a searching query $q$, and then the Retriever returns a long list of passages $p_s$ based on the similarity search within the given Document corpus $D$. Based on the retrieved information and the initial question, Refiner selects the most relevant topk passages as $p_r$. IM storage is now updated with the searching query and $p_r$. 
Following the above-described working flow of \textit{Progress Tracker}, Line~\ref{alg:line-PT}~to~\ref{alg:line-PT-end} conclude One Round of IM by calculating the distance score $d$ and update the $SP$ list. 
This multi-round process continues until the \textit{Progress Tracker} determines that the $SP$ is empty. 
After all necessary information has been gathered, to complete the IM process, the \textit{Questioner} will also provide the final prediction $A_f$. In the open-format QA task, we consider both $A_f$ and ground-truth answer $G$ as a sequence of tokens. Thus, as shown in Equation~\ref{eq:major-reward}, the precision and recall of the predicted answer can be used to calculate the F1 score.

\begin{equation}
r = F1(A_{f}, G)
\label{eq:major-reward}
\end{equation}

From $i$th-round of Inner Monologue, \textit{Progress Tracker} collects $i$ number of distance scores. As part of the final reward, $1 - d_i$ is used 
to reflect the quality of $i$-th round of retrieval in a continuous space.
We introduce a discount factor, $\gamma < 1$, to emphasize the importance of the preceding search. 
Inheriting from IMMO, the reward also includes the KL divergence with a predefined weight, $\alpha$, between the updated \textit{Questioner} $\mathcal{M}$ and its starting point $\mathcal{M}_0$~\citep{jaques2017sequence, ziegler2019fine}. 
The final reward is a non-discrete number, which depends on both the IM quality (distance score) and the answer quality (correctness score). The \textit{Questioner} LoRA is updated by the PPO algorithm driven by the reward function as shown in Equation~\ref{eq:final-reward}.

\begin{equation}
    \mathcal{R} = ( \sum_{i=1}^{n} \gamma^i (1 - d_i ) ) + r - \alpha KL(\mathcal{M}, \mathcal{M}_0)
\label{eq:final-reward}
\end{equation}

\subsubsection{Answerer Training}
\label{sec:answerer}
After the \textit{Questioner} has been trained, it learned the ability to perform a reasonable IM, thus obtaining valid supporting evidence from the IR module. As discussed, the goal of asking meaningful questions differs from final question answering. Thus, we define an \textit{Answerer}, which specializes in the QA capability to be exclusively responsible for providing the final answer. 

In most datasets or tasks, the final answers are provided, and the multi-round retrieval (IM) information can be acquired by the well-trained \textit{Questioner}. Therefore, we have sufficient data to support supervised learning. Following the instruction fine-tuning technique~\citep{chung2022scaling, alpaca}, the training data can be prepared as a combination of the Initial Question, Inner Monologue, and Final Answer. The training object for \textit{Answerer} Lora is to perform the next token prediction over the corpus. 

\section{Experiment}
In this section, we introduce the task and data in the experiment, the implementation and training details of our IM-RAG approach, the baseline approaches we compared with, and the experiment results verified with statistical significance.

\subsection{Task and Data}
IM-RAG targets the multi-hop retrieval question-answering task. In this kind of task, the knowledge needed to solve the problem usually exists in multiple passages from a given document corpus. For the experiment, we test IM-RAG in HotPotQA, which is a widely used open-domain multi-hop QA dataset. 

HotPotQA involves providing a system with a set of related documents and a question that requires reasoning across these documents to arrive at an answer. The input consists of the question and the list of supporting documents, while the output is the answer to the question, which can be in the form spanning from text from the documents, a yes/no response, to a sentence. 
Additionally, HotPotQA provides a document corpus that includes all introductory paragraphs from English Wikipedia 2017. The task is to identify the supporting facts within the document corpus that led to the answer. 
We follow the original data split to conduct the experiment and report the result on the dev set following the community convention on this dataset. The evaluation is done by the official script from HotPotQA, which includes EM (Exact Matching) and F1 score between the predicted answer and the ground-truth answer label. Besides, since the related supporting documents are provided as a list, the retrieval result can also be evaluated by EM and F1. This setup encourages the development of models that are not only adept at extracting answers but also capable of understanding the context and performing multi-hop reasoning. As our system is designed for final task completion, we focus more on the evaluation of the final answer.

\subsection{Implementation Details}
\label{sec:impl_details}
Below, we provide the implementation details of IM-RAG, which follows the approach design illustrated in Section~\ref{sec:approach}.

\subsubsection{Reasoner}
\label{sec:impl_reasoner}
Following the design from Section~\ref{sec:reasoner}, we utilize a large pretrained language model as the \textit{Reasoner} in IM-RAG. Specifically, we use the 7B version of Vicuna-1.5~\citep{vicuna2023} as the base LLM, which is an open-source LLM fine-tuned from LLaMA-2~\citep{touvron2023llama} with supervised instruction fine-tuning on 125K high-quality user-shared conversations collected from ShareGPT~\citep{shareGPT}. Building upon the base LLM, we add and finetune two LoRA adapters as the \textit{Questioner} and the \textit{Answerer}, respectively. As discussed in Section~\ref{sec:reasoner}, this design allows the capabilities of \textit{Questioner} and the \textit{Answerer} to be separately learned while fully reusing the same base LLM.

\begin{table*}[tb]
\centering
\resizebox{0.75\linewidth}{!}{
  \setlength{\tabcolsep}{3pt}
  \begin{tabular}{l c c c c c c}
    \toprule
     {Method} & Multi-rounds & RAG~\tablefootnote{The RAG categorization follows our definition in Section~\ref{sec:related}.}
     & Training & Passage EM & EM & F1 \\
    \midrule
     GPT-3.5 & No & LLM-centric & Prompt &  N/A & 31.0 & 37.1 \\
     REACT~\citep{yao2022react} & Yes & LLM-centric & Prompt & - & 35.1 & - \\
    \midrule
     TPRR~\citep{trrr} & Yes & Jointly Train & SFT & 86.2 & 67.3 & 80.1 \\
     AISO~\citep{zhu2021adaptive} & Yes & Jointly Train & RL & 88.2 & 68.1 & 80.9 \\
     COS~\citep{ma-etal-2023-chain} & Yes & Jointly Train & SFT & 88.9 & 68.2 & 81.0 \\
    \midrule
     RAG (no IM) & No & LLM-centric & SFT & 36.2 & 31.2 & 41.2 \\
     IM-RAG & Yes & LLM-centric & RL+SFT & 83.4 & 68.4 & 82.5\\
    \bottomrule
  \end{tabular}
  }
  \caption{Results on HotPotQA. The results were categorized into three groups based on training data and the type of RAG paradigm.}
 \label{tab:results}
\end{table*}

\subsubsection{Retriever}
\label{sec:impl_retriever}
Following the Dense Passage Retrieval (DPR) approach~\citep{karpukhin2020dense}, we index 5.2 million supporting documents using Sentence-transformer~\citep{reimers-2019-sentence-bert} embedding, which is fine-tuned for semantic search on a question-to-document matching task. We use FAISS library~\citep{johnson2019billion} to facilitate rapid similarity searches, averaging 0.061 seconds per query under the GPU environment. Due to the flexibility of our approach, the \textit{Retriever} can be replaced with stronger search engines or fine-tuned to further boost the IR performance, while based on the experiments on the HotPotQA dataset, our current \textit{Retriever} setting has already met the accuracy, speed, and scalability requirements by our approach.

\subsubsection{Refiner}
\label{sec:impl_refiner}
Given the experimental design where the output from \textit{Retriever} is a list of Wikipedia introductory paragraphs retrieved by FAISS from HotPotQA, the primary goal of \textit{Refiner} is to rerank this list, prioritizing the supporting facts. Given the effectiveness and rapid deployability of LLM-reranker, as demonstrated in previous works \citep{sun2023chatgpt, pradeep2023rankzephyr}, we employ the checkpoint of RankVicuna \citep{pradeep2023rankvicuna}, an LLM pretrained for listwise document reranking. The reasons for selecting RankVicuna are as follows: (1) As a pre-trained LLM, RankVicuna allows us to effortlessly harness its language comprehension and zero-shot capabilities for ranking tasks across various documents, eliminating the need for additional fine-tuning. (2)~\citet{ke2024bridging} highlighted a significant gap between retrievers and LLMs, which often impedes their communication, and proposed to add a seq2seq model to enhance the output of retrievers. We found that RankVicuna, as a variant of the fine-tuned Vicuna LLMs, matches the size and base capabilities of the \textit{Reasoner} (also a Vicuna LLM), effectively bridging the gap and facilitating the overall IM process.

\subsubsection{Progress Tracker}
\label{sec:impl_progress_tracker}
As discussed in section~\ref{sec:approach}, the design of the \textit{Progress Tracker} can be flexible across different tasks. In HotPotQA, as the ground-truth supporting documents are provided, we implemented the \textit{Progress Tracker} in a heuristic way. Specifically, given the list of ground-truth documents $SP$ and retrieved document $p_i$, we compute the cosine similarity between $p_i$ with each element in $SP$ in the Sentence-transformer embedding space. The distance to the closest one will be recorded as the distance score $d_i$ for the training as described in section~\ref{sec:approach}. Moreover, this document will be considered as retrieved, so it will be removed from $SP$ and will not be involved in the next-turn comparison. This design provides dependencies across IM turns and encourages the \textit{Reasoner} to search for new documents. In addition to the $SP$ status mentioned in \textit{Questioner} training (Section~\ref{sec:questioner}), the switch between the \textit{Questioner} and the \textit{Answerer} is also controlled by an empirically selected threshold $\phi_r$ for the accumulated distance reward scores $\mathcal{D}$ over multiple turns as well as a preset maximum number of turns $N_{max}$ (see Equation~\ref{eq:distance_score}~and~\ref{eq:switch_qa_func}). If $\mathcal{D}$ is below the threshold $\phi_r$, the \textit{Reasoner} will continue the responsibility of the \textit{Questioner} to craft a new query for retrieval. Conversely, as enough information has been collected or $N_{max}$ has been reached, the \textit{Reasoner} will switch to the \textit{Answerer} to provide a final answer to the question. In the experiment, we set $\phi_r$ to $0.3$ and $N_{max}$ to $3$.

\begin{equation}
\mathcal{D} = \sum_{i=1}^{n} \gamma^i (1 - d_i )
\label{eq:distance_score}
\end{equation}

\begin{equation}
Reasoner = 
\begin{cases} 
  Questioner, & \text{if } \mathcal{D} \leq \phi_r  \text{ and } i < N_{max} \\
  Answerer, & \text{if } \mathcal{D} > \phi_r \text{ or } i = N_{max}
\end{cases}
\label{eq:switch_qa_func}
\end{equation}

\subsection{Training Details}
\label{sec:training_details}
Following the previous works~\citep{yang2023tackling, alpaca, accelerate}, the RL of \textit{Questioner} is supported by Transformers Reinforcement-Learning (TRL) library~\citep{vonwerra2022trl}, and the SFT of \textit{Answerer} is supported by the HuggingFace instruction finetuning pipeline~\citep{wolf-etal-2020-transformers}. All the hyperparameters follow the default settings from StackLLaMA~\citep{beeching2023stackllama} and Alpaca~\citep{alpaca}. With the Parameter-Efficient Fine-Tuning (PEFT)~\citep{peft} support, under a 4 {NVIDIA} A100 GPU environment, the \textit{Questioner} (RL) and \textit{Answerer} (SFT) are trained for 6 and 10 epochs, respectively. The instruction prompt is modified from the template provided by previous works~\citep{yang2023tackling, alpaca}.

\subsection{Baselines}
\label{sec:baseline}

We compared IM-RAG with three groups of baseline approaches. 
The first group relies on the power of LLM and can be plug-and-play by other available similar models or APIs. GPT-3.5 delivers QA results without connecting to an external knowledge base. We provide 4-shot in-context examples as instruction for the LLM. REACT~\citep{yao2022react}, as one of the early RAG works, chains LLMs with search engines via prompting and in-context examples. It is a simple yet effective approach with good zero-shot performance.

We also include several good-performing, representative works in the HotPotQA dataset. It is important to note that our focus is on the enhancement of the LLM-centric system rather than developing a comprehensive QA system. The inclusion of these works primarily serves as a reference for performance.
AISO~\citep{zhu2021adaptive} models the QA task as a Reinforcement Learning trained Markov decision process (MDP), whose action space includes the selection of different retrieval algorithms and the answer generation. This sophisticated system achieves promising results; however, it is expensive to adapt this training-from-scratch system to a new domain. Instead of a complex MDP, IM-RAG uses LLM as the policy network, so it can be easily optimized for a new domain by policy-based learning method~\citep{ouyang2022training, stiennon2020learning}. 
Another noteworthy work is Chain-of-Skill (CoS)~\citep{ma-etal-2023-chain}, which employs manually designed domain-specific retrieval skills (such as entity linking and expanded query retrieval, etc.) for Q\&A tasks. These carefully designed skills significantly improve the performance of language models; however, domain knowledge may required to design the new skills when adapting to a new domain. Specifically, CoS learns how to use skills through a multi-task pre-training phase, which needs to be retrained for a new domain or skills change. AISO also has a similar challenge.
In addition, both AISO and CoS are inherently tied to predefined IR systems. This means that plug-and-play other custom search modules or knowledge bases are not straightforward. In general, both approaches heavily rely on domain expertise for system design and require retraining when design changes.

The last baseline, RAG (no IM), shares a similar structure with IM-RAG as well as the modeling selection; the only difference is that it does not support multi-round retrieval due to the absence of the IM process. This baseline uses the initial question as the retrieval query to obtain the documents that will be needed for supervised training for the \textit{Answerer}. 

\subsection{Results}
The results are reported in Table~\ref{tab:results}. Compared to the prompting-based approach, IM-RAG gains significant improvements while retaining flexibility. Previous work pointed out that ChatGPT falls short in ensuring factuality in complex QA problems~\citep{zheng2023does}. In our comparison, GPT3.5 lagged behind RAG (no IM) by 0.2\% and 4.1\% on EM and F1 scores, respectively. REACT, powered by PaLM-540B~\citep{palm}, shows strong zero-shot capability; however, due to the limited task-specific optimization, it does not have the advantage in terms of performance compared to the approaches with training. 

Compared to the second group of works that are usually tied to predefined IR systems, IM-RAG has better flexibility in IR module selection. In our comparison, IM-RAG outperformed the previous best-performed model by 1.9\% relative gain on F1 score. On the other hand, IM-RAG lagged behind others in the second group in retrieval metrics like Passage EM because our focus wasn't on fine-tuning the IR module. However, LLM's rich pre-training knowledge tolerates imperfect retrieval information and overturns the final QA result.

For the last baseline, with the same model selection and system design, IM-RAG outperforms the RAG (no IM) baseline by a huge margin (82.5\% vs. 41.2\%) in terms of F1 score. We claim that the multi-round retrieval is the key to the success of the IM-RAG framework. 

\paragraph{Significance Test}

\begin{table}[h]
\centering
\label{table:mcnemar}
\begin{tabular}{lcc}
\hline
\textbf{Model Comparison} & \textbf{p-Value} & \textbf{Significance} \\ 
\hline
IM-RAG vs. no-IM & $< 0.001$ & Yes \\
IM-RAG vs. GPT-3.5  & $< 0.001$ & Yes \\ 
IM-RAG vs. no-SFT & $0.008$ & Yes \\
IM-RAG vs. no-Refiner & $< 0.001$ & Yes \\ 
\hline
\end{tabular}
\caption{McNemar test results for comparing IM-RAG with other LLM-based methods. All test shows the IM-RAG result is statistically significant.}
\label{tab:test}
\end{table}

In this study, we employed McNemar's test~\citep{McNemar1947} using Statsmodels~\citep{seabold2010statsmodels} to statistically evaluate the performance improvements of our IM-RAG model compared to two baselines approaches mentioned in Section~\ref{sec:baseline} (no-IM and GPT-3) and two results from ablation study (no-SFT and no-Refiner) on HotPotQA~\footnote{Limited by available resources, we were unable to obtain prediction files of other baselines. Therefore, we performed significance tests only for the above methods.}. The test is conducted on the prediction following the EM (0, 1) measurement. This non-parametric test is particularly suited for binary labels on paired nominal data. As reported in Table~\ref{tab:test}, the test results indicated that the IM-RAG model demonstrated a statistically significant improvement in performance over all the above-mentioned approaches. 

\section{Ablation Study and Analysis}
In this section, we conduct an ablation study to investigate and analyze how different training strategies and components impact the performance of IM-RAG, as well as outline the limitations of IM-RAG.

\begin{table}[tb]
\resizebox{\linewidth}{!}{
  \centering
  \setlength{\tabcolsep}{3pt}
  \begin{tabular}{|c|c|c|c|c|}
    \hline
     Questioner (RL) & Answerer (SFT) & Refiner & EM & F1 \\
    \hline
    \color{red}{\ding{55}} & \color{teal}{\ding{51}} & \color{teal}{\ding{51}} & Error & Error \\
    \color{teal}{\ding{51}} & \color{red}{\ding{55}} & \color{teal}{\ding{51}} & 63.9 & 77.9 \\
    \color{teal}{\ding{51}} & \color{teal}{\ding{51}} & \color{red}{\ding{55}} & 35.5 & 48.3 \\
    \color{teal}{\ding{51}} & \color{teal}{\ding{51}} & \color{teal}{\ding{51}} & 68.4 & 82.5 \\
    \hline
  \end{tabular}
  }
  \caption{Ablation Study on each component in IM-RAG. Error indicates the system fails to work under the given setting. }
 \label{tab:ablation}\vspace{-3mm}
\end{table}

\subsection{The Impact of Training Strategy}

The complete training process of IM-RAG includes reinforcement learning as well as supervised learning. Thus, we report two ablation experiments in this section to reveal the respective impacts.
As shown in table~\ref{tab:ablation}, first, we remove the RL training for \textit{Questioner}. The plan is to enable the LLM to engage the multi-round retrieval by prompting and in-context examples. This approach can be regarded as ``prompting the Inner Monologue". After collecting the query and the retrieved documents, we train the \textit{Answerer} Lora in the same way as mentioned in Section~\ref{sec:answerer}. However, in our experiments, we were unable to control the LLM (vicuna-7b) to output in the desired format. Under the zero-shot scenario, for a large number of data points, the LLM generates irrelevant content or does not provide the query. Potential solutions would be to use a more powerful language model (e.g., GPT-4 or LlaMA2-70b) or a more sophisticated prompt design. However, the former requires huge computational resources, whereas the latter requires more effort from humans.

Another set of experiments focused on the effects of supervised fine-tuning. As shown in Algorithm~\ref{alg:RL}, since the \textit{Questioner} training originally includes providing final prediction, we can simply remove the \textit{Answerer} LoRA and record the \textit{Questioner}'s response after completing the retrieval as the prediction. Under the same experimental configuration, the \textit{Questioner} LoRA obtained 77.9\% F1 score. There is a 4.6\% decrease from 82.5\% (full version IM-RAG). As explained in section~\ref{sec:approach}, asking for supporting facts and answering based on retrieved information require two different abilities. Assigning the tasks to two models (or two LoRAs in our design) simplifies the challenge, resulting in improved performance.

\subsection{Necessity of the Refiner}

As discussed in Section~\ref{sec:approach}, the purpose of the \textit{Refiner} is to improve the output of the \textit{Retriever}, which effectively bridges the gap between the \textit{Reasoner} and the \textit{Retriever}, and fosters the IM process. To better understand the necessity of the \textit{Refiner}, we conduct an ablation study to explore how the \textit{Refiner} impacts the performance of IM-RAG. In the experiment design on HotPotQA, the \textit{Refiner} plays the role of a re-ranker to highlight the most relevant passages. As a comparison, we run another experiment where we simply use the top-5 passages provided by the \textit{Retriever} at each turn without involving the \textit{Refiner} for further refinement.

As shown in Table~\ref{tab:ablation}, with all other settings consistent, removing the \textit{Refiner} leads to a 14.2\% performance drop (68.3\% vs. 82.5\%) in terms of the F1 score. This result can be attributed to the gap between the IR module and the LLM~\citep{ke2024bridging}. As introduced in Section~\ref{sec:approach}, in the process of learning IM, the \textit{Reasoner} actively proposes queries at each turn to acquire more relevant documents from the \textit{Retriever}. However, there exists a gap between the \textit{Reasoner} and the \textit{Retriever}, specifically in the format, length, and importance of the retrieved documents compared to the expected context for the \textit{Reasoner}. Such a gap may not only give the \textit{Reasoner} a "hard time" in figuring out the most relevant information from the retrieved documents, but also hinder the \textit{Progress Tracker} from giving a positive reward that guides the IM learning via RL. In the cases where a large training corpus exists, the \textit{Reasoner} might be able to learn how to fill the gap through intensive training, while this is more costly and less efficient. Therefore, we can conclude that the \textit{Refiner} is a necessary component to bridge the gap and facilitate IM learning.

\section{Discussion}
This section discusses situations in which IM-RAG applies as well as those in which it does not.

\paragraph{Task}
IM-RAG benefits from the rich language ability of the pretrained LLM and excels in capturing dynamic information and then performing context-aware multi-round retrieval. Thus, it specializes in multi-hop retrieval and generation tasks. However, the performance of IM-RAG in single-step accurate retrieval and real-world complex environments is unclear. 
\paragraph{IR Dependency}
The mobilized design makes IM-RAG very easy to be applied to customization tasks. Depending on the retrieval scenario or domain, the IR module in Figure~\ref{fig:overview} can be replaced by other wildly-designed search engines or dense retrievers. 
\paragraph{Data Requirement} For migration on a new task, the most challenging aspect is the preparation and acquisition of the data required by the \textit{Progress Tracker}. During training, the retrieval quality signals provided by \textit{Progress Tracker} directly guide the optimization of the strategy. In our experiments, \textit{Progress Tracker} used the ground-truth retrieval results provided by the training set. However, in cases where more resources are available (e.g., search logs from real users), \textit{Progress Tracker} can provide better guidance for the training of the \textit{Reasoner}. In contrast, when the available resources are unable to support \textit{Progress Tracker} to provide retrieval score, IM-RAG will be stuck in the massive language (action) space and thus unable to optimize because it can hardly reach the positive reward.
\paragraph{Inference Efficiency}
Similar to other LLM-based RAG work~\citep{sun2023chatgpt, izacard2022few}, in general, IM-RAG has the higher inference latency than traditional IR systems~\citep{10.1145/1076034.1076103, 10.1145/1835449.1835511}. As a result, it is difficult for IM-RAG to meet the speed requirement in contexts where it is necessary to obtain a fast response, and conversely, LLM brings decent reasoning ability as well as generative results.

\section{Limitation and Future Works}
This work demonstrates promising results in utilizing Inner Monologue to solve traditional information retrieval tasks; however, the potential of the IM-RAG framework has not been fully explored. 
As discussed above, an important advantage of this framework is the reinforcement of the model's reasoning ability through outcome supervision. Compared to employing supervised learning to impart models to do Chain-of-Thought reasoning, this approach facilitates models to find superior solutions, i.e., the reasoning path that is better suited to their own system capabilities. 
However, due to the RL's optimization difficulties on language models, this work uses final result supervision along with another strong reward signal, i.e., the human-labeled golden document is considered as the target answer for each round of retrieval. 
This signaling serves as a fine guide during training yet sets an upper limit to IM retrieval. We expect that this problem can be solved in the future by better \textit{Progress Tracker} design, such as pretraining a complex neural network to provide retrieval signals directly without the supervision of the golden documents from humans. Following the idea of RLHF~\citep{ouyang2022training}, using a large number of human annotations to train a reward model to act as a \textit{Progress Tracker} is a promising approach. However, this design may only be available to institutions with the resources to do so.

\section{Conclusion}
We present IM-RAG, a novel approach inspired by inner monologues, which connects LLM and IR to accomplish complex reasoning tasks through context-aware multi-round interactive retrieval. During multi-round conversations, the LLM serves as the core reasoning model, either crafting new queries for the retriever based on the conversational context or generating a final response when enough information has been collected. The retrieved documents are modified (reformatted, re-ranked, filtered, etc.) by the \textit{refiner} to better match the needs of LLM. The whole process can be optimized end-to-end via RL using the feedback from the \textit{Progress Tracker} and final answer correctness as reward signals. The results on HotPotQA show that IM-RAG achieves SOTA performance in multi-step reasoning. This enables the RAG system to do human-like multi-round reasoning and retrieval with high flexibility and interpretability.

While this is the first step towards learning how to do inner monologue between LLM and retrievers, as with all preliminary research, it comes with certain limitations. The dataset we used may not reflect the subtle and sometimes non-linear nature of human inner monologue, potentially limiting the model's ability to learn and handle highly complex, abstract, or creative reasoning tasks.

\bibliographystyle{ACM-Reference-Format}
\bibliography{main-camera-ready}

\appendix

\end{document}